# Best-First AND/OR Search for Most Probable Explanations

**Radu Marinescu** and **Rina Dechter**
School of Information and Computer Science
University of California, Irvine, CA 92697-3425
{radum,dechter}@ics.uci.edu

## Abstract

The paper evaluates the power of *best-first search* over AND/OR search spaces for solving the Most Probable Explanation (MPE) task in Bayesian networks. The main virtue of the AND/OR representation of the search space is its sensitivity to the structure of the problem, which can translate into significant time savings. In recent years depth-first AND/OR Branch-and-Bound algorithms were shown to be very effective when exploring such search spaces, especially when using caching. Since best-first strategies are known to be superior to depth-first when memory is utilized, exploring the best-first control strategy is called for. The main contribution of this paper is in showing that a recent extension of AND/OR search algorithms from depth-first Branch-and-Bound to best-first is indeed very effective for computing the MPE in Bayesian networks. We demonstrate empirically the superiority of the best-first search approach on various probabilistic networks.

## 1 INTRODUCTION

Belief networks [1] are a class of graphical models that provide a formalism for reasoning about partial beliefs under conditions of uncertainty. They are defined by a directed acyclic graph over nodes representing random variables of interest. The arcs signify the existence of direct causal influences between linked variables quantified by conditional probabilities that are attached to each cluster of parents-child nodes in the network. The Most Probable Explanation (MPE) task in belief networks calls for finding a complete assignment to the variables having maximum probability, given the evidence. It is typically tackled with either *inference* or *search* algorithms [1, 2, 3].

The AND/OR search space for graphical models [4] is a framework for search that is sensitive to the independencies in the model, often resulting in reduced search spaces. The impact of the AND/OR search to optimization in graphical models and in particular to the MPE task was explored in recent years focusing exclusively on depth-first search.

The AND/OR Branch-and-Bound first introduced by [3] traverses the AND/OR search tree in a depth-first manner. The memory intensive Branch-and-Bound algorithm [5] explores an AND/OR search graph, rather than a tree, by caching previously computed results and retrieving them when the same subproblems are encountered again. The depth-first AND/OR search algorithms were shown to outperform dramatically state-of-the-art Branch-and-Bound algorithms searching the traditional OR space.

In a recent paper [6] we introduced best-first AND/OR search algorithms for solving 0-1 Integer Programming problems, and demonstrated that, given enough memory, they are superior to Branch-and-Bound algorithms we developed earlier [7]. Subsequently, in [8] we extended this approach for Weighted CSP (WCSP) problems when using best-first AND/OR search guided by bounded mini-bucket heuristics. We demonstrated, again, that the best-first algorithms are more efficient than their Branch-and-Bound counterparts for various hard WCSP benchmarks.

In this paper we shift our attention to probabilistic networks, focusing on the MPE tasks. The extension of best-first AND/OR search from WCSP to Bayesian networks is straightforward. Hence, the main contribution of the current paper is in its empirical evaluation of the scheme over a wide range of probabilistic networks, including random networks, coding networks as well as hard instances from genetic linkage analysis. We show that this class of algorithms improves on the most competitive complete MPE solvers, thus it can potentially push the landmark of computation further, assuming memory is available.

The paper is organized as follows. Section 2 gives background on belief networks and AND/OR search spaces. Section 3 describes the best-first AND/OR search algorithm. Section 4 presents an extensive empirical evaluation and Section 5 concludes.

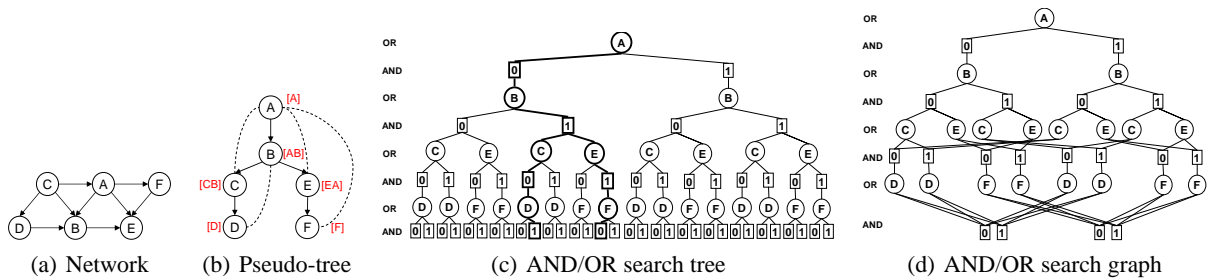

Figure 1: AND/OR search spaces for belief networks.

## 2 BACKGROUND

### 2.1 Belief Networks

DEFINITION **1 (belief network)** *A* belief (or Bayesian) network *is a quadruple $\mathcal{P} = \langle X, D, F \rangle$, where $X = \{X_1, ..., X_n\}$ is a set of variables over multi-valued domains $D = \{D_1, ..., D_n\}$. Given a directed acyclic graph $DAG$ over $X$ as nodes, $F = \{P_i\}$, where $P_i = \{P(X_i|pa(X_i))\}$ are conditional probability tables (CPTs for short) associated with each variable $X_i$, and $pa(X_i)$ are the parents of $X_i$ in the acyclic graph $DAG$. A belief network represents a joint probability distribution over $X$, $P(x_1, ..., x_n) = \prod_{i=1}^{n} P(x_i|x_{pa(X_i)})$. An evidence set $e$ is an instantiated subset of variables. The* moral graph (or primal graph) *of a belief network is the undirected graph obtained by connecting the parent nodes of each variable and eliminating direction.*

A common optimization query over belief networks is finding the *Most Probable Explanation* (MPE), namely, finding a complete assignment to all variables having maximum probability, given the evidence. A generalization of the MPE query is *Maximum a Posteriori Hypothesis* (MAP), which calls for finding the most likely assignment to a subset of hypothesis variables, given the evidence.

DEFINITION **2 (most probable explanation)** *Given a belief network and evidence $e$, the* Most Probable Explanation (MPE) *task is to find an assignment $(x_1^o, ..., x_n^o)$ such that: $P(x_1^o, ..., x_n^o) = max_{X_1,...,X_n} \prod_{k=1}^{n} P(X_k|pa(X_k), e)$.*

The MPE task appears in applications such as diagnosis, abduction and explanation. For example, given data on clinical findings, MPE can postulate on a patient's probable afflictions. In decoding, the task is to identify the most likely message transmitted over a noisy channel given the observed output.

### 2.2 AND/OR Search Spaces for Graphical Models

The common way to solve the MPE task in belief networks is by search, namely to instantiate variables, following a static or dynamic variable ordering. In the simplest case, this process defines an OR search tree, whose nodes represent partial assignments. This search space does not capture the structure of the underlying graphical model. However, to remedy this problem, AND/OR search spaces for graphical models were recently introduced by [4]. They are defined using a backbone *pseudo-tree* [9].

DEFINITION **3 (pseudo-tree)** *Given an undirected graph $G = (V, E)$, a directed rooted tree $T = (V, E')$ defined on all its nodes is called* pseudo-tree *if any arc of $G$ which is not included in $E'$ is a back-arc, namely it connects a node to an ancestor in $T$.*

**AND/OR Search Trees**   Given a belief network $\mathcal{P} = \langle X, D, F \rangle$, its primal graph $G$ and a pseudo-tree $T$ of $G$, the associated AND/OR search tree, denoted $S_T$, has alternating levels of OR nodes and AND nodes. The OR nodes are labeled $X_i$ and correspond to the variables. The AND nodes are labeled $\langle X_i, x_i \rangle$ and correspond to value assignments in the domains of the variables. The root of the AND/OR search tree is an OR node, labeled with the root of the pseudo-tree $T$.

The children of an OR node $X_i$ are AND nodes labeled with assignments $\langle X_i, x_i \rangle$, consistent along the path from the root, $path(X_i, x_i) = (\langle X_1, x_1 \rangle, ..., \langle X_{i-1}, x_{i-1} \rangle)$. The children of an AND node $\langle X_i, x_i \rangle$ are OR nodes labeled with the children of variable $X_i$ in $T$. Semantically, the OR states represent alternative solutions, whereas the AND states represent problem decomposition into independent subproblems, all of which need be solved. When the pseudo-tree is a chain, the AND/OR search tree coincides with the regular OR search tree.

A *solution tree* $Sol_{S_T}$ of $S_T$ is an AND/OR subtree such that: (i) it contains the root of $S_T$; (ii) if a nonterminal AND node $n \in S_T$ is in $Sol_{S_T}$ then all its children are in $Sol_{S_T}$; (iii) if a nonterminal OR node $n \in S_T$ is in $Sol_{S_T}$ then exactly one of its children is in $Sol_{S_T}$.

EXAMPLE **1** *Figures 1(a) and 1(b) show a belief network and its pseudo-tree together with the back-arcs (dotted lines). Figure 1(c) shows the AND/OR search tree based on the pseudo-tree, for bi-valued variables.*

**Weighted AND/OR Search Trees**   The arcs from OR nodes $X_i$ to AND nodes $\langle X_i, x_i \rangle$ in the AND/OR search tree $S_T$ are annotated by *weights* derived from the conditional probability tables in $F$.

DEFINITION **4 (weight)** *The* weight $w(n, m)$ *of the arc from the OR node* $n = X_i$ *to the AND node* $m = \langle X_i, x_i \rangle$ *is the product of all the CPTs whose scope includes* $X_i$ *and is fully assigned along* $path(X_i, x_i)$, *evaluated at the values along the path.*

Given a weighted AND/OR search tree, each node can be associated with a *value* [4].

DEFINITION **5 (value)** *The* value $v(n)$ *of a node* $n \in S_T$ *is defined recursively as follows: (i) if* $n = \langle X_i, x_i \rangle$ *is a terminal AND node then* $v(n) = 1$; *(ii) if* $n = \langle X_i, x_i \rangle$ *is an internal AND node then* $v(n) = \prod_{m \in succ(n)} v(m)$; *(iii) if* $n = X_i$ *is an internal OR node then* $v(n) = max_{m \in succ(n)}(w(n, m) \cdot v(m))$, *where* $succ(n)$ *are the children of* $n$ *in* $S_T$.

It easy to see that the value $v(n)$ of a node in the AND/OR search tree $S_T$ is the most probable explanation of the subproblem rooted at $n$, subject to the current variable instantiation along the path from the root to $n$. If $n$ is the root of $S_T$, then $v(n)$ is the most probable explanation value of the initial problem (see [3, 4] for more details).

**AND/OR Search Graphs**  The AND/OR search tree may contain nodes that root identical subtrees (in particular, subproblems with identical optimal solutions) which can be *unified*. When unifiable nodes are merged, the search tree becomes a graph and its size becomes smaller. Some unifiable nodes can be identified based on their *contexts*.

DEFINITION **6 (context)** *Given a belief network and the corresponding AND/OR search tree* $S_T$ *relative to a pseudo-tree* $T$, *the* context *of any AND node* $\langle X_i, x_i \rangle \in S_T$, *denoted by* $context(X_i)$, *is defined as the set of ancestors of* $X_i$ *in* $T$, *including* $X_i$, *that are connected to descendants of* $X_i$.

It is easy to verify that any two nodes having the same context represent the same subproblem. Therefore, we can solve $P_{\langle X_i, x_i \rangle}$, the subproblem rooted at $\langle X_i, x_i \rangle$, once and use its optimal solution whenever the same subproblem is encountered again.

The *context-minimal* AND/OR search graph based on a pseudo-tree $T$, denoted $G_T$, is obtained from the AND/OR search tree by merging all the AND nodes that have the same context. It can be shown [4] that the size of the largest context is bounded by the induced width $w^*$ of the problem's primal graph.

THEOREM **2.1 (complexity)** *The complexity of any search algorithm traversing a context-minimal AND/OR search graph (by context-based caching) is time and space* $O(exp(w^*))$, *where* $w^*$ *is the induced width of the underlying pseudo-tree [4].*

EXAMPLE **2** *Consider the context-minimal AND/OR search graph in Figure 1(d) of the pseudo-tree from Figure 1(b) (the square brackets indicate the context of the variables). Its size is far smaller than that of the AND/OR tree from Figure 1(c) (16 nodes vs. 36 nodes).*

### 2.3  Searching the AND/OR Search Space

Recently, depth-first AND/OR Branch-and-Bound (AOBB) search algorithms that explore the context-minimal AND/OR search graph via full caching were shown to be highly effective for solving the MPE task in belief networks [3, 5]. The efficiency of these algorithms also depends on the accuracy of a *static heuristic function* which can be either pre-compiled or generated during search for each node in the search space. Furthermore, we showed [3] that AOBB can improve its guiding heuristic function dynamically, by learning from portions of the search space that were already explored. This updated *dynamic heuristic evaluation function* is guaranteed to be tighter than the static one [3], and therefore it can prune the search space more effectively. The primary static heuristic function we experimented with, especially in the context of the MPE task was the *mini-bucket* heuristic [2].

**The Mini-bucket Heuristics** is a general scheme for generating heuristic estimates for search that has been investigated in recent years, especially in the context of belief networks [2, 3, 5]. The scheme is parameterized by the mini-bucket $i$-bound which controls the trade-off between heuristic strength and its computational overhead. The heuristics can be pre-compiled from the augmented bucket structure processed by the Mini-Bucket algorithm. When compiled before search they are referred to as *static mini-buckets* (hereafter denoted by SMB) and they were shown to be very powerful, especially for relatively large values of the $i$-bound. When the mini-bucket heuristics are computed dynamically during search, referred to as *dynamic mini-buckets* (DMB) they are generally more accurate than the static ones. However, due to their computational overhead, they were shown to be cost effective only for relatively small $i$-bounds.

## 3  BEST-FIRST AND/OR SEARCH

In this section we direct our attention to a *best-first* rather than depth-first control strategy for traversing the context-minimal AND/OR graph and describe a best-first AND/OR search algorithm for solving the MPE task in belief networks. The algorithm uses similar amounts of memory as the depth-first AND/OR Branch-and-Bound with full caching and therefore the comparison is warranted.

---

**Algorithm 1**: AOBF

**Data**: A belief network $\mathcal{P} = \langle X, D, F \rangle$, pseudo-tree $T$, root $s$.
**Result**: Most Probable Explanation of $\mathcal{P}$.

1. Create explicit graph $G'_T$, consisting solely of the start node $s$. Set $v(s) = h(s)$.
2. **until** $s$ is labeled SOLVED, **do**:
   (a) Compute a *partial solution tree* by tracing down the *marked* arcs in $G'_T$ from $s$ and select any nonterminal tip node $n$.
   (b) Expand node $n$ and add any new successor node $n_i$ to $G'_T$. For each new node $n_i$ set $v(n_i) = h(n_i)$. Label SOLVED any of these successors that are terminal nodes.
   (c) Create a set $S$ containing node $n$.
   (d) **until** $S$ is empty, **do**:
      i. Remove from $S$ a node $m$ such that $m$ has no descendants in $G'_T$ still in $S$.
      ii. Revise the value $v(m)$ as follows:
         A. **if** $m$ is an AND node **then**
         $v(m) = \prod_{m_j \in succ(m)} v(m_j)$. If all the successor nodes are labeled SOLVED, then label node $m$ SOLVED.
         B. **if** $m$ is an OR node **then**
         $v(m) = max_{m_j \in succ(m)}(w(m, m_j) \cdot v(m_j))$ and mark the arc through which this maximum is achieved. If the marked successor is labeled SOLVED, then label $m$ SOLVED.
      iii. If $m$ has been marked SOLVED or if the revised value $v(m)$ is different than the previous one, then add to $S$ all those parents of $m$ such that $m$ is one of their successors through a marked arc.
3. **return** $v(s)$.

---

**Best-First Search** Best-first search is a search algorithm which optimizes breath-first search by expanding the node whose heuristic evaluation function is the best among all nodes encountered so far. Its main virtue is that it never expands nodes whose cost is beyond the optimal one, unlike depth-first search algorithms, and therefore is superior among memory intensive algorithms employing the same heuristic evaluation function [10].

**Best-First AND/OR Graph Search** Our best-first AND/OR graph search algorithm, denoted by AOBF, that traverses the context-minimal AND/OR search graph is described in Algorithm 1. It specializes Nilsson's AO* algorithm [11] to AND/OR spaces in graphical models, in particular to finding the MPE in belief networks.

The algorithm maintains a frontier of partial solution trees found so far, and interleaves forward expansion of the best partial solution tree with a cost revision step that updates estimated node values. First, a top-down, graph-growing operation (step 2.a) finds the best partial solution tree by tracing down through the marked arcs of the explicit AND/OR search graph $G'_T$. These previously computed marks indicate the current best partial solution tree from each node in $G'_T$. One of the nonterminal leaf nodes $n$ of this best partial solution tree is then expanded, and a static heuristic estimate $h(n_i)$, overestimating $v(n_i)$, is assigned to its successors (step 2.b). The successors of an AND node $n = \langle X_j, x_j \rangle$ are $X_j$'s children in the pseudo-tree, while the successors of an OR node $n = X_j$ correspond to $X_j$'s domain values. Notice that when expanding an OR node, the algorithm does not generate AND children that are already present in the explicit search graph $G'_T$. All these identical AND nodes in $G'_T$ are easily recognized based on their contexts, so only pointers to the existing nodes are created.

The second operation in AOBF is a bottom-up, cost revision, arc marking, SOLVE-labeling procedure (step 2.c). Starting with the node just expanded $n$, the procedure revises its value $v(n)$ (using the newly computed values of its successors) and marks the outgoing arcs on the estimated best path to terminal nodes. This revised value is then propagated upwards in the graph. The revised cost $v(n)$ is an updated estimate of the most probable explanation probability of the subproblem rooted at $n$. If we assume the monotone restriction on $h$, the algorithm considers only those ancestors that root best partial solution subtrees containing descendants with revised values. The most probable explanation value of the initial problem is obtained when the root node $s$ is solved.

**AOBF versus AOBB** We describe next the main differences between AOBF and AOBB search.

1. AOBF with the same heuristic function as AOBB is likely to expand the smallest number of nodes [10], but empirically this depends on how quickly AOBB will find an optimal solution.

2. AOBB is able to improve its heuristic function dynamically during search [3] based on the explicated portion of the search space, while AOBF may not because it uses only the static function $h(n)$, which can be precomputed or generated during search.

3. AOBB can use far less memory avoiding dead-caches for example (e.g., when the search graph is a tree), while AOBF has to keep the explicated search graph in memory prior to termination.

All the above points show that the relative merit of best-first vs depth-first over context-minimal AND/OR search spaces cannot be determined by the theory in [10] and empirical evaluation is essential.

## 4 EXPERIMENTS

We evaluate the performance of the best-first AND/OR search algorithm on the task of finding the Most Probable Explanation in belief networks [1]. We implemented our algorithms in C++ and ran all experiments on a 2.4GHz Pentium IV with 2GB of RAM.

We consider a class of best-first AND/OR search algorithms guided by the static and dynamic mini-bucket heuristics. They are denoted by AOBF+SMB($i$) and AOBF+DMB($i$), respectively. We compare them against the depth-first AND/OR Branch-and-Bound algorithms with static/dynamic mini-bucket heuristics and full caching introduced in [5] and denoted by AOBB+SMB($i$) and AOBB+DMB($i$) respectively. The parameter $i$ represents the mini-bucket $i$-bound and controls the accuracy of the heuristic. All algorithms traverse the context-minimal AND/OR search graph and are restricted to a static variable ordering determined by the pseudo-tree. In our current implementation the AND/OR search algorithms do not exploit the determinism present in the networks by using any form of constraint propagation such as generalized arc-consistency or unit propagation.

For reference, we include results obtained with the SAMIAM 2.3.2 software package[1]. SAMIAM is a public implementation of Recursive Conditioning [13] which can also be viewed as an AND/OR search algorithm.

We report the average CPU time in seconds (t) and number of nodes visited (#), required for proving optimality of the solution. We also record the number of variables (n), number of evidence variables (e), the depth of the pseudo-trees (h) and the induced width of the graphs ($w^*$) obtained for the test instances. The pseudo-trees were generated using the min-fill heuristic, as described in [3]. All competing algorithms were alloted a 2GB memory limit. The best performance points are highlighted. In each table, "-/out" denotes that the respective algorithm exceeded the time/memory limit.

**Random Belief Networks** We have generated a class of random belief networks using the parametric model $(n, d, c, p)$ proposed in [2]. Figure 2 reports the average time results in seconds and number of nodes visited for 20 random instances of a network with $n = 120$ variables, domain size $d = 2$, $c = 110$ probability tables (CPTs) and $p = 2$ parents per CPT. The average induced width and pseudo-tree depth were 20 and 32, respectively. The mini-bucket $i$-bound ranged between 2 and 16.

When comparing the best-first versus the depth-first algorithms using static mini-bucket heuristics, we observe that AOBF+SMB($i$) is better than AOBB+SMB($i$) only for relatively small $i$-bounds (i.e., $i \in \{2, 3, 4\}$) which generate relatively weak heuristic estimates. As the $i$-bound increases and the heuristics become strong enough to cut the search space substantially, the difference between Branch-and-Bound and best-first search decreases, because Branch-and-Bound finds close to optimal solutions fast, and therefore will not explore solutions whose cost is below the optimum, like best-first search.

When looking at the algorithms using dynamic mini-bucket heuristics, we notice that AOBF+DMB($i$) is slightly bet-ter than AOBB+DMB($i$) only for the smallest reported $i$-bound, namely $i = 2$. This is because these heuristics are more accurate compared to the static ones, and the savings in number of nodes caused by best-first search do not transform into time savings as well. When comparing the static versus dynamic mini-bucket heuristic we observe that the latter is competitive only for relatively small $i$-bounds (i.e., $i \in \{2, 3, 4, 5, 6\}$). At higher levels of the $i$-bound, the accuracy of the dynamic heuristic does not outweigh its computational overhead. For this reason, in the remaining experiments we only consider the algorithms guided by pre-compiled mini-bucket heuristics.

[1] Available at http://reasoning.cs.ucla.edu/samiam. We used the batchtool 1.5 provided with the package.

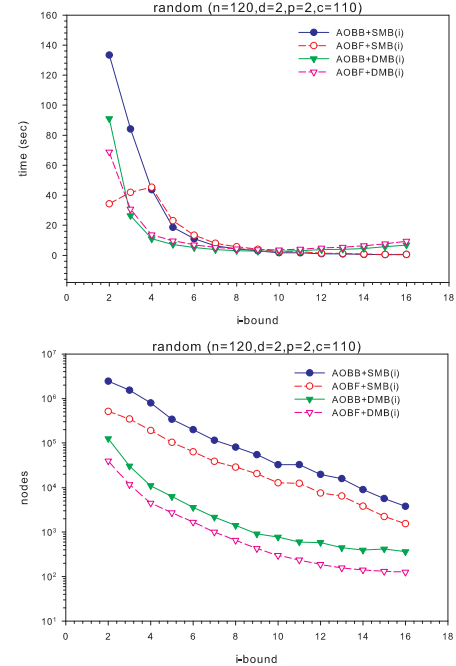

Figure 2: CPU time in seconds and number of nodes visited for solving random belief networks with 120 nodes. Time limit 180 seconds, average induced width $w^* = 20$.

**Coding Networks** For this domain we experimented with random coding networks from the class of *linear block codes*. They can be represented as 4-layer belief networks with $n$ nodes in each layer (i.e., the number of input bits). The second and third layers correspond to input information bits and parity check bits respectively. Each parity check bit represents an XOR function of the input bits. The first and last layers correspond to transmitted information and parity check bits respectively. Input information and parity check nodes are binary, while the output nodes are real-valued. Given a number of input bits $n$, number of parents $p$ for each XOR bit, and channel noise variance $\sigma^2$, a coding network structure is generated by randomly picking parents for each XOR node. Then we simulate an input signal by assuming a uniform random distribution of information bits, compute the corresponding values of the parity check bits, and generate an assignment to the output nodes by adding Gaussian noise to each information and parity check bit.

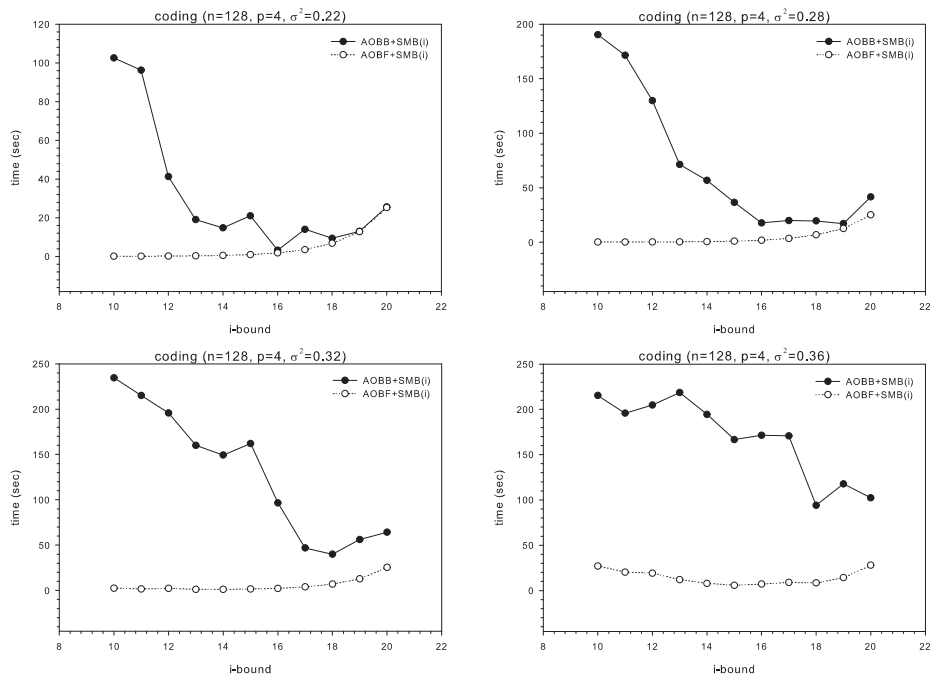

Figure 3: CPU time in seconds for solving coding networks with channel noise variance $\sigma^2 \in \{0.22, 0.28, 0.32, 0.36\}$. Time limit 300 seconds, average induced width $w^* = 54$.

Figure 3 displays the average time results in seconds for 20 random coding instances with $n = 128$ input bits, $p = 4$ parents for each XOR bit and channel noise variance $\sigma^2 \in \{0.22, 0.28, 0.32, 0.36\}$ (we omitted the number of nodes due to space limitations). The average induced width and depth of the pseudo-tree was 54 and 71, respectively. The mini-bucket $i$-bound varied between 10 and 20. We observe that AOBF+SMB($i$) is far better than AOBB+SMB($i$) for this domain. The difference in CPU time between the best-first and depth-first search approaches is more prominent on the hardest problem instances having higher channel noise variance (i.e., $\sigma^2 \in \{0.32, 0.36\}$), across all reported $i$-bounds. SAMIAM was not able to solve any of these problems due to exceeding the memory limit.

**Grid Networks** In grid networks, the nodes are arranged in an $n \times n$ square and each CPT is generated uniformly randomly. We experimented with problem instances developed by [14] for which $n$ ranged between 10 and 38, and 90% of the CPTs were deterministic (i.e., constraints).

Table 1 shows detailed results for experiments with 8 grid networks of increasing difficulty. For each network $e$ nodes were picked randomly and instantiated as evidence. We notice again the superiority of AOBF+SMB($i$) over AOBB+SMB($i$), especially for relatively weak heuristic estimates which are generated at relatively small $i$-bounds.

For example, on 90-34-1, one of the hardest instances, best-first search with the smallest reported $i$-bound ($i = 12$) finds the most probable explanation in about 8 minutes (495 seconds) while the depth-first Branch-and-Bound with the same heuristics exceeds the 1 hour time limit. The best performance point on this test instance is achieved for $i = 18$, where AOBF+SMB(18) is 9 times faster than AOBB+SMB(18) and explores a search space 23 times smaller. Notice that SAMIAM is able to solve relatively efficiently only the first 3 test instances and runs out of memory on the remaining ones.

**Genetic Linkage Analysis** The *maximum likelihood haplotype* problem in genetic linkage analysis is the task of finding a joint haplotype configuration for all members of the pedigree which maximizes the probability of data. It is equivalent to finding the most probable explanation of a belief network which represents the pedigree data [15].

Table 2 displays the results obtained for 12 hard linkage analysis networks[2]. For comparison, we include results obtained with SUPERLINK 1.6. SUPERLINK is currently one the most efficient solvers for genetic linkage analysis, is dedicated to this domain, uses a combination of variable elimination and conditioning, and takes advantage of the determinism in the network.

We observe again that AOBF+SMB($i$) is the best performing algorithm. For instance, on the p42 linkage instance, AOBF+SMB(14) is 18 times faster than AOBB+SMB(14)

---
[2]http://bioinfo.cs.technion.ac.il/superlink/

| grid | n e | w* h | | SamIam v. 2.3.2 | AOBB+SMB(i) | | | | | AOBF+SMB(i) | | | | |
|---|---|---|---|---|---|---|---|---|---|---|---|---|---|---|
| | | | | | i=8 | i=10 | i=12 | i=14 | i=16 | i=8 | i=10 | i=12 | i=14 | i=16 |
| 90-10-1 | 100 | 16 | t | 0.13 | 0.23 | 0.19 | **0.08** | 0.11 | 0.19 | 0.22 | 0.14 | **0.08** | 0.09 | 0.19 |
| | 0 | 26 | # | | 4,396 | 3,681 | 1,231 | 760 | 101 | 1,788 | 1,046 | 517 | 312 | 100 |
| 90-14-1 | 196 | 23 | t | 11.97 | 19.95 | 12.52 | 8.83 | 1.22 | 0.78 | 8.24 | 5.97 | 2.20 | 1.02 | **0.70** |
| | 0 | 37 | # | | 215,723 | 156,387 | 112,962 | 14,842 | 4,209 | 46,153 | 35,537 | 13,990 | 5,137 | 1,163 |
| 90-16-1 | 256 | 26 | t | 147.19 | 1223.55 | 130.47 | 11.09 | 11.25 | **2.38** | 133.19 | 47.72 | 9.91 | 10.53 | 2.97 |
| | 0 | 42 | # | | 13,511,366 | 1,469,593 | 135,746 | 123,841 | 18,230 | 673,238 | 250,098 | 55,112 | 52,644 | 11,854 |
| | | | | | i=12 | i=14 | i=16 | i=18 | i=20 | i=12 | i=14 | i=16 | i=18 | i=20 |
| 90-24-1 | 576 | 36 | t | out | 1237.19 | 285.63 | 75.02 | 22.83 | 20.78 | 34.21 | 38.35 | 13.49 | **9.08** | 21.00 |
| | 20 | 61 | # | | 6,922,516 | 2,051,503 | 547,401 | 110,144 | 15,400 | 125,962 | 149,445 | 49,261 | 14,390 | 8,155 |
| 90-26-1 | 676 | 35 | t | out | - | - | 634.59 | 85.11 | 49.97 | out | out | 57.66 | **29.08** | 32.95 |
| | 40 | 64 | # | | | | 4,254,454 | 455,404 | 169,942 | | | 190,527 | 66,429 | 24,487 |
| 90-30-1 | 900 | 38 | t | out | - | - | 365.69 | 145.86 | 37.39 | out | out | 40.80 | 40.67 | **36.00** |
| | 60 | 68 | # | | | | 2,837,671 | 936,463 | 32,637 | | | 136,576 | 121,561 | 13,217 |
| 90-34-1 | 1154 | 43 | t | out | - | - | 974.65 | 534.10 | 522.05 | 494.69 | 175.85 | 88.24 | **59.39** | 90.19 |
| | 80 | 79 | # | | | | 5,555,182 | 2,647,012 | 2,430,599 | 705,922 | 303,782 | 189,340 | 112,955 | 115,553 |
| 90-38-1 | 1444 | 47 | t | out | - | 81.27 | 657.91 | 734.46 | 133.06 | 478.02 | **22.80** | 47.14 | 43.74 | 78.05 |
| | 120 | 86 | # | | | 259,405 | 1,505,849 | 1,478,903 | 161,156 | 580,623 | 38,376 | 80,177 | 52,209 | 35,294 |

Table 1: CPU time in seconds and number of nodes visited for solving grid networks. Time limit 1 hour.

| ped | n | w* h | | SamIam v. 2.3.2 | Superlink v. 1.6 | AOBB+SMB(i) | | | | | AOBF+SMB(i) | | | | |
|---|---|---|---|---|---|---|---|---|---|---|---|---|---|---|---|
| | | | | | | i=6 | i=8 | i=10 | i=12 | i=14 | i=6 | i=8 | i=10 | i=12 | i=14 |
| p1 | 299 | 15 | t | 5.44 | 54.73 | 4.19 | 2.17 | 0.39 | 0.65 | 1.36 | 1.30 | 2.17 | **0.26** | 0.87 | 1.54 |
| | | 61 | # | | | 69,751 | 33,908 | 4,576 | 6,306 | 4,494 | 7,314 | 13,784 | 1,177 | 4,016 | 3,119 |
| p38 | 582 | 17 | t | out | **28.36** | 5946.44 | 1554.65 | 2046.95 | 272.69 | | out | 134.41 | 216.94 | 103.17 | |
| | | 59 | # | | | 34,828,046 | 8,986,648 | 11,868,672 | 1,412,976 | | | 348,723 | 583,401 | 242,429 | |
| p50 | 479 | 18 | t | out | - | 4140.29 | 2493.75 | 66.66 | 52.11 | | 78.53 | 36.03 | **12.75** | 38.52 | |
| | | 58 | # | | | 28,201,843 | 15,729,294 | 403,234 | 110,302 | | 204,886 | 104,289 | 25,507 | 5,766 | |
| | | | | | | i=10 | i=12 | i=14 | i=16 | i=18 | i=10 | i=12 | i=14 | i=16 | i=18 |
| p23 | 310 | 23 | t | out | 9146.19 | 53.70 | 49.33 | 8.77 | **2.73** | 3.04 | 35.49 | 29.29 | 10.59 | 3.59 | 3.48 |
| | | 37 | # | | | 486,991 | 437,688 | 85,721 | 14,019 | 7,089 | 185,761 | 150,214 | 52,710 | 11,414 | 5,790 |
| p37 | 1032 | 21 | t | out | 64.17 | 39.16 | 488.34 | 301.78 | 67.83 | | **29.16** | 38.41 | 95.27 | 62.97 | |
| | | 61 | # | | | 222,747 | 4,925,737 | 2,798,044 | 82,239 | | 72,868 | 102,011 | 223,398 | 12,296 | |
| | | | | | | i=12 | i=14 | i=16 | i=18 | i=20 | i=12 | i=14 | i=16 | i=18 | i=20 |
| p18 | 1184 | 21 | t | 157.05 | 139.06 | - | 406.88 | 52.91 | 23.83 | 20.60 | out | 127.41 | 42.19 | **19.85** | 19.91 |
| | | 119 | # | | | | 3,567,729 | 397,934 | 118,869 | 2,972 | | 542,156 | 171,039 | 53,961 | 2,027 |
| p20 | 388 | 23 | t | out | **14.72** | 7243.43 | 5560.63 | 37.28 | 95.13 | | out | out | 33.33 | 121.91 | |
| | | 42 | # | | | 63,530,037 | 46,858,127 | 279,804 | 554,623 | | | | 144,212 | 466,817 | |
| p25 | 994 | 29 | t | out | - | - | - | - | 2041.64 | 693.74 | out | out | out | out | **198.49** |
| | | 53 | # | | | | | | 6,117,320 | 1,925,152 | | | | | 468,723 |
| p30 | 1016 | 25 | t | out | 13095.83 | 1440.26 | 597.88 | 1023.90 | 151.96 | 43.83 | 186.77 | 58.38 | 85.53 | 49.38 | **33.03** |
| | | 51 | # | | | 11,694,534 | 5,580,555 | 10,458,174 | 1,179,236 | 146,896 | 692,870 | 253,465 | 350,497 | 179,790 | 37,705 |
| p33 | 581 | 26 | t | out | - | 886.05 | 370.41 | 26.31 | 33.11 | 54.89 | out | 194.78 | **24.16** | 32.55 | 58.52 |
| | | 48 | # | | | 8,426,659 | 4,032,864 | 229,856 | 219,047 | 83,360 | | 975,617 | 102,888 | 101,862 | 57,593 |
| p39 | 1272 | 23 | t | out | 322.14 | - | - | 968.03 | 61.20 | 93.19 | out | out | 68.52 | **41.69** | 87.63 |
| | | 94 | # | | | | | 7,880,928 | 313,496 | 83,714 | | | 218,925 | 79,356 | 14,479 |
| p42 | 448 | 25 | t | out | 561.31 | - | - | 2364.67 | | | out | out | **133.19** | | |
| | | 76 | # | | | | | 22,595,247 | | | | | 93,831 | | |

Table 2: CPU time in seconds and number of nodes visited for genetic linkage analysis. Time limit 3 hours.

| bn | n | w* h | | SamIam v. 2.3.2 | AOBB+SMB(i) | | | | | AOBF+SMB(i) | | | | |
|---|---|---|---|---|---|---|---|---|---|---|---|---|---|---|
| | | | | | i=16 | i=18 | i=20 | i=21 | i=22 | i=16 | i=18 | i=20 | i=21 | i=22 |
| BN_031 | 1153 | 46 | t | out | 1183.49 | 541.82 | 217.80 | 83.08 | 145.55 | 187.95 | 125.94 | 83.89 | **71.53** | 132.55 |
| | | 160 | # | | 3,990,212 | 2,131,977 | 889,782 | 94,507 | 97,721 | 427,788 | 292,293 | 114,046 | 25,392 | 30,067 |
| BN_033 | 1441 | 43 | t | - | 1717.53 | 157.17 | 190.77 | 129.74 | 154.16 | 80.58 | **41.25** | 73.70 | 94.52 | 143.58 |
| | | 163 | # | | 2,156,432 | 210,552 | 256,191 | 89,308 | 46,312 | 124,453 | 41,865 | 49,760 | 22,256 | 14,894 |
| BN_035 | 1441 | 41 | t | - | 67.74 | 133.28 | 58.81 | 80.64 | 157.83 | **27.25** | 36.75 | 51.20 | 75.53 | 158.17 |
| | | 168 | # | | 174,370 | 243,533 | 65,657 | 58,973 | 45,758 | 31,460 | 34,987 | 15,953 | 18,048 | 18,461 |
| BN_037 | 1441 | 45 | t | - | 34.77 | 21.28 | 45.20 | 90.35 | 144.60 | **12.80** | 19.25 | 45.88 | 90.30 | 146.61 |
| | | 169 | # | | 69,326 | 33,475 | 8,815 | 16,400 | 12,507 | 16,304 | 11,046 | 4,315 | 5,610 | 4,798 |
| BN_039 | 1441 | 48 | t | - | 1727.89 | 475.26 | 246.60 | 653.83 | | out | 254.25 | 113.97 | **112.69** | 211.84 |
| | | 162 | # | | 3,448,072 | 1,043,378 | 518,011 | 3,045,139 | | | 725,738 | 213,676 | 127,872 | 239,838 |
| BN_041 | 1441 | 49 | t | - | 257.96 | 56.66 | 54.36 | 78.74 | 130.94 | 36.22 | **22.20** | 43.56 | 69.91 | 121.24 |
| | | 164 | # | | 354,822 | 77,653 | 38,467 | 31,763 | 38,088 | 94,220 | 20,485 | 16,549 | 11,648 | 16,533 |
| BN_127 | 512 | 57 | t | out | 1798.57 | - | - | 128.55 | 113.06 | **54.03** | 58.84 | 64.53 | 66.34 | 121.53 |
| | | 74 | # | | 17,583,748 | | | 860,026 | 93,543 | 235,416 | 251,134 | 166,741 | 84,007 | 70,351 |
| BN_129 | 512 | 52 | t | out | 640.29 | - | 1439.32 | 222.17 | 155.63 | out | 200.47 | **135.60** | out | 231.95 |
| | | 68 | # | | 6,150,175 | | 13,437,762 | 1,747,613 | 671,931 | | 922,831 | 537,371 | | 622,449 |
| BN_131 | 512 | 48 | t | out | - | 43.06 | 51.16 | - | 156.11 | **19.67** | 50.58 | 36.66 | 65.75 | 99.20 |
| | | 72 | # | | | 396,234 | 303,818 | | 759,649 | 82,780 | 209,748 | 73,163 | 120,153 | 46,662 |
| BN_134 | 512 | 52 | t | out | - | - | - | - | 234.38 | out | **86.80** | 96.21 | 97.28 | 112.63 |
| | | 70 | # | | | | | | 1,438,986 | | 373,081 | 377,064 | 214,591 | 102,530 |

Table 3: CPU time in seconds and number of nodes visited for solving UAI'06. Time limit 30 minutes.

and explores a search space 240 times smaller. On some instances (e.g., p1, p23, p30) the best-first search algorithm AOBF+SMB($i$) is several orders of magnitude faster than SUPERLINK. The performance of SAMIAM was very poor on this dataset and it was able to solve only 2 instances.

**UAI'06 Evaluation Dataset**   We also experimented with 10 belief networks from the UAI'06 Evaluation Dataset[3]. We were not able to obtain the code from the other competitors (*i.e.*, Teams 1 and 2) in the MPE evaluation, and therefore we only compare against AOBB and SAMIAM.

Table 3 displays a summary of the results. We observe that AOBF+SMB($i$) is the best performing algorithm on this dataset. While on the first 6 instances AOBF+SMB($i$) improves only slightly causing on average a 2.5 speed-up over AOBB+SMB($i$), on the remaining 4 instances, the difference between best-first and depth-first search is more dramatic. For example, AOBF+SMB(18) solves the BN_134 instance in less than 2 minutes, while AOBB+SMB(18) exceeds the 30 minute time limit. We notice that in some cases (e.g. BN_127, BN_129), especially for large mini-bucket $i$-bounds (e.g. $i=22$) which generate very accurate heuristic estimates, the savings in number of nodes caused by AOBF+SMB($i$) do not outweigh its overhead.

**Summary of experiments.**   In summary, best-first AND/OR search with static/dynamic mini-bucket heuristics improves dramatically over depth-first AND/OR Branch-and-Bound search, especially for relatively weak heuristic estimates which are generated for relatively small mini-bucket $i$-bounds. This is significant because it allows the best-first search algorithms to push the landmark of computation further as the induced width of the problems increases.

## 5   CONCLUSION

In this paper we evaluated a best-first AND/OR search algorithm which extends the classic AO* algorithm and traverses a context-minimal AND/OR search graph for solving the MPE task in belief networks. The algorithm is guided by mini-bucket heuristics which can be either pre-compiled or assembled dynamically during search. The efficiency of the best-first AND/OR search approach compared to the depth-first AND/OR Branch-and-Bound search is demonstrated empirically on various random and real-world benchmarks, including the very challenging ones that arise in the field of genetic linkage analysis.

Our approach leaves room for further improvements. The space required by AOBF can be enormous, due to the fact that all the nodes generated by the algorithm have to be saved prior to termination. Therefore, AOBF can be extended to incorporate a memory bounding scheme similar to the one suggested in [16].

---

[3] http://ssli.ee.washington.edu/bilmes/uai06InferenceEvaluation

**Acknowledgments**

This work was supported by the NSF grant IIS-0412854.

**References**

[1] J. Pearl. *Probabilistic Reasoning in Intelligent Systems.* Morgan-Kaufmann, 1988.

[2] K. Kask and R. Dechter. A general scheme for automatic generation of search heuristics from specification dependencies. *Artificial Intelligence*, 2001.

[3] R. Marinescu and R. Dechter. And/or branch-and-bound for graphical models. *In IJCAI*, pages 224–229, 2005.

[4] R. Dechter and R. Mateescu. And/or search spaces for graphical models. *Artificial Intelligence*, 2006.

[5] R. Marinescu and R. Dechter. Memory intensive branch-and-bound search for graphical models. *In AAAI*, 2006.

[6] R. Marinescu and R. Dechter. Best-first and/or search for 0-1 integer programming. *In CPAIOR*, 2007.

[7] R. Marinescu and R. Dechter. And/or branch-and-bound search for pure 0/1 integer linear programming problems. *In CPAIOR*, pages 152–166, 2006.

[8] R. Marinescu and R. Dechter. Best-first and/or search for graphical models. *In AAAI*, 2007.

[9] E. Freuder and M. Quinn. Taking advantage of stable sets of variables in constraint satisfaction problems. *In IJCAI*, pages 1076–1078, 1985.

[10] R. Dechter and J. Pearl. Generalized best-first search strategies and the optimality of a*. *In Journal of ACM*, 32(3):505–536, 1985.

[11] Nils J. Nilsson. *Principles of Artificial Intelligence.* Tioga, 1980.

[12] R. Dechter and I. Rish. Mini-buckets: A general scheme for approximating inference. *ACM*, 2003.

[13] A. Darwiche. Recursive conditioning. *Artificial Intelligence*, 126(1-2):5–41, 2001.

[14] T. Sang, P. Beame, and H. Kautz. Solving Bayesian networks by weighted model counting. *In AAAI*, pages 475–482, 2005.

[15] M. Fishelson, N. Dovgolevsky, and D. Geiger. Maximum likelihood haplotyping for general pedigrees. *Human Heredity*, 2005.

[16] P. Chakrabati, S. Ghose, A. Acharya, and S. de Sarkar. Heuristic search in restricted memory. *In Artificial Intelligence*, 3(41):197–221, 1989.